\documentclass{article}
\usepackage[latin9]{inputenc}
\usepackage{array}
\usepackage{float}
\usepackage{units}
\usepackage{multirow}
\usepackage{amsmath}
\usepackage{graphicx}

\makeatletter

\providecommand{\tabularnewline}{\\}

\usepackage{amssymb}
\usepackage{spconf}
\usepackage{epsfig}
\usepackage{epstopdf}
\usepackage{graphics}
\ninept
\let\textquotedbl="

\title{Study of Set-Membership Kernel Adaptive Algorithms and
Applications
 \vspace{-0.15em}}
 \name{Andre Flores and Rodrigo C. de Lamare,Senior Member, IEEE
\vspace{-0.75em}}
\address{\\
    CETUC,PUC-Rio,Brazil \\
    Email: andre.flores@cetuc.puc-rio.br, delamare@cetuc.puc-rio.br\vspace{-0.35em}
    }

\makeatother

\begin{document}
\maketitle
\begin{abstract}
Adaptive algorithms based on kernel structures have been a topic of
significant research over the past few years. The main advantage is
that they form a family of universal approximators, offering an elegant
solution to problems with nonlinearities. Nevertheless these methods
deal with kernel expansions, creating a growing structure also known
as dictionary, whose size depends on the number of new inputs. In
this paper we derive the set-membership kernel-based normalized least-mean
square (SM-NKLMS) algorithm, which is capable of limiting the size
of the dictionary created in stationary environments. We also derive
as an extension the set-membership kernelized affine projection (SM-KAP)
algorithm. Finally several experiments are presented to compare the
proposed SM-NKLMS and SM-KAP algorithms to the existing methods. \vspace{-0.25em}

\end{abstract}
\begin{keywords} Kernel methods, sparsification, set-membership kernel
adaptive filtering. \end{keywords}\vspace{-0.5em}

\section{Introduction}

Adaptive filtering algorithms have been the focus of a great deal
of research in the past decades, and the machine learning community
has embraced and further advanced the study of these methods. However,
conventional adaptive algorithms often work with linear strucutres,
limiting the performance that they can achieve and constraining the
number of problems that can be solved. Under this scope a new family
of nonlinear adaptive filter algorithms based on kernels was developed.
A kernel is a function that compares the similarity between two inputs.
The kernel adaptive filtering (KAF) algorithms have been tested in
many different scenarios and applications \cite{Gil-Cacho2013}\cite{Gil-Cacho2012}
\cite{Nakijama2012}\cite{RichardBermudezHoneine2009},
showing very good results.

As described in \cite{LiuPrincipeHaykin2010}, one of the main advantages
of KAF algorithms is that they are universal approximators, which
gives them the capability to treat complex and nonlinear problems.
In other words, they can model any input-output mapping. Many of these
algorithms have no local minima, which is also a desirable characteristic.
However, the computational complexity is significantly higher than
their linear counterparts\cite{Theodoridis2015}.

One of the first KAF algorithms to appear and widely adopted in the
KAF family because of its simplicity is the kernel least-mean square
(KLMS) proposed in \cite{LiuPokharelPrincipe2008} and extended in
\cite{Boboulis2011}. The KLMS algorithm is inspired by the least-mean
square algorithm and showed good results, so that many researchers
have worked since then in the development of new kernel versions of
conventional adaptive algorithms. A few years later, a kernelized
version of the NLMS algorithm was proposed in \cite{RichardBermudezHoneine2009}
using a nonlinear regression approach for time series prediction.
In \cite{LiuPrincipe2008}, the affine projection algorithm (APA)
was modified to develop a family of four algorithms known as the kernel
affine projection algorithms (KAPA). The recursive least squares algorithm
(RLS) was also extended in \cite{Engel2004}, where the kernel recursive
least squares was introduced (KRLS). Later , the authors of \cite{Liu2009}
proposed an extended version of the KRLS algorithm. Also the use of
multiple kernels was studied in \cite{Pokharel2013} and \cite{Yukawa2012}.

All the algorithms mentioned before have to deal with kernel expansions.
In other words, they create a growing structure, also called dictionary,
where they keep every new data input that arrives to compute the estimate
of the desired output. The natural problem that arises is that the
time and computational cost spent to compute a certain output could
exceed the tolerable limits for a specific application. Several criteria
were proposed to solve this problem. One of the most simple criteria
is the novelty criterion, presented in \cite{Platt1991}. Basically
it establishes two thresholds to limit the size of the dictionary.
Another method, the approximate linear dependency (ALD) was proposed
in \cite{Engel2004} and verifies if a new input can be expressed
as a linear combination of the elements stored before adding this
input to the dictionary. The coherence criterion was introduced in
\cite{RichardBermudezHoneine2009} also to limit the size of the dictionary
based on the similarity of the inputs. A measure called surprise was
presented in \cite{Liu2009a} to remove redundant data.

In this work, we present the set-membership normalized kernel
least-mean square (SM-NKLMS) and the set-membership kernel affine
projection (SM-KAP) adaptive algorithms, which can provide a faster
learning than existing kernel-based algorithms and limit the size of
the dictionary without compromising performance. Similarly to
existing set-membership algorithms
\cite{Lamare2009,DinizWerner2003,GollamudiNagarajKapoor1998,smtvb,smce,smjio,sm-ccm,smbf,smcg,WernerDiniz2001},
the proposed SM-NKLMS and SM-KAP algorithms are equipped with
variable step sizes and perform sparse updates that are useful for
several applications
\cite{aifir,jiolms,smstnr,ccmmwf,wlmwf,wlrrbf,ccg,jiols,rccm,jiomimo,jidf,sjidf,gsc-radar,jio-radar,sa-stap,mserjidf,zu2014multi,zhang2014robust,locsme,arh,als,jidfdoa,okspme,dce,damdc}.
Unlike existing kernel-based adaptive algorithms the proposed
SM-NKLMS and SM-KAP algorithms deal with in a natural way with the
kernel expansion because of the data selectivity based on error
bounds that they implement.

This paper is organized as follows. In section II, the problem formulation
is presented. In section III the SM-NKLMS and the SM-KAP algorithms
are derived. Section IV presents the simulations and results of the
algorithms developed in an application involving a time series prediction
task. Finally, section V presents the conclusions of this work. \vspace{-0.05em}

\section{Problem Statement}

\noindent Let us consider an adaptive filtering problem with a sequence
of training samples given by $\left\{ \boldsymbol{x}\left[i\right],d\left[i\right]\right\} $,
where $\boldsymbol{x}\left[i\right]$ is the N-dimensional input vector
of the system and $d[i]$ represents the desired signal at time instant
$i$. The output of the adaptive filter is given by

\begin{equation}
y\left[i\right]=\mathbf{w}^{T}\boldsymbol{x}\left[i\right],
\end{equation}
where $\mathbf{w}$ is the weight vector with length N.

Let us define a non-linear transformation denoted by $\boldsymbol{\varphi:}\mathbb{R}\rightarrow\mathbb{F}$
that maps the input to a high-dimensional feature space. Applying
the transformation stated before, we map the input and the weights
to a high-dimensional space obtaining:
\begin{equation}
\boldsymbol{\varphi}\left[i\right]=\boldsymbol{\varphi}\left(\boldsymbol{x}\left[i\right]\right),
\end{equation}
\begin{equation}
\boldsymbol{\omega}\left[i\right]=\boldsymbol{\varphi}(\mathbf{w}\left[i\right]),
\end{equation}

The error generated by the system is given by $e\left[i\right]=d\left[i\right]-\boldsymbol{\omega}^{T}\left[i\right]\boldsymbol{\varphi}\left[i\right]$.
The main objective of the kernel-based adaptive algorithms is to model
a function to implement an input-output mapping, such that the mean
square error generated by the system is minimized. In addition, we
assume that the magnitude of the estimated error is upper bounded
by a quantity $\gamma$. The idea of using an error bound was reported
in \cite{FogelHuang1982} and was used since then to develop different
versions of data selective algorithms.

\vspace{-0.35em}

\section{Proposed Set-Membership Kernel-Based Algorithms}

\vspace{-0.85em}
Assuming that the value of $\gamma$ is correctly chosen then there
exists several functions that satisfy the error requirement. To summarize,
any function leading to an estimation error smaller than the defined
threshold is an adequate solution, resulting in a set of filters.
Consider a set $\boldsymbol{\bar{S}}$ containing all the possible
input-desired pairs of interest $\left\{ \boldsymbol{\varphi}\left[i\right],d\left[i\right]\right\} $.
Now we can define a set $\boldsymbol{\theta}$ with all the possible
functions leading to an estimation error bounded in magnitude by $\gamma$.
This set is known as the feasibility set and is expressed by

\begin{equation}
\boldsymbol{\theta}=\bigcap_{\left\{ \boldsymbol{\varphi},d\right\} \in\boldsymbol{\bar{S}}}\left\{ \boldsymbol{\omega}\in\mathbb{F}\:/\:|d-\boldsymbol{\omega}^{T}\boldsymbol{\varphi}|\leq\gamma\right\}
\end{equation}

Suppose that we are only interested in the case in which only measured
data are available. Let us define a new set $\mathcal{H}\left[i\right]$
with all the functions such that the estimation error is upper bounded
by $\gamma$ . The set is called constraint set and is mathematically
defined by
\begin{equation}
\mathcal{H}\left[i\right]=\left\{ \boldsymbol{\omega}\in\mathbb{F}\:/\:|d\left[i\right]-\boldsymbol{\omega}^{T}\boldsymbol{\varphi}\left[i\right]|\leq\gamma\right\}
\end{equation}

It follows that for each data pair there exists an associated constraint
set. The set containing the intersection of the constraint sets over
all available time instants is called exact membership set and is
given by the following equation:
\begin{equation}
\psi\left[i\right]=\bigcap_{k=0}^{i}\mathcal{H}\left[i\right]
\end{equation}

The exact membership set,$\psi\left[i\right]$, should become small
as the data containing new information arrives. This means that at
some point the adaptive filter will reach a state where $\psi\left[i\right]=\psi\left[i-1\right]$,
so that there is no need to update $\boldsymbol{\omega\left[i\right]}$.
This happens because $\psi\left[i-1\right]$ is already a subset of
$\mathcal{H}\left[i\right]$.As a result, the update of any set-membership
based algorithm is data dependent, saving resources, a fact that is
crucial in kernel-based adaptive filters because of the growing structure
that they create.

As a first step we check if the previous estimate is outside the constraint
set, i.e., $|d\left[i\right]-\boldsymbol{\omega}^{T}\left[i-1\right]\boldsymbol{\varphi}\left[i\right]|>\gamma$.
If the error exceeds the bound established, the algorithm performs
an update so that the a posteriori estimated error lies in $\mathcal{H}\left[i\right]$
.If the previous case occurs we minimize $||\boldsymbol{\omega}\left[i+1\right]-\boldsymbol{\omega}\left[i\right]||^{2}$
subject to $\boldsymbol{\omega}\left[i+1\right]\in\mathcal{H}\left[i\right]$,
which means that the a posteriori error $\xi_{ap}\left[i\right]$
is given by

\begin{equation}
\xi_{ap}\left[i\right]=d\left[i\right]-\boldsymbol{\omega}^{T}\left[i+1\right]\boldsymbol{\varphi}\left[i\right]=\pm\gamma\label{eq:posteriori error}
\end{equation}

The NKLMS update equation presented in \cite{LiuPrincipeHaykin2010}
is given by:
\begin{equation}
\boldsymbol{\omega}\left[i+1\right]=\boldsymbol{\omega}\left[i\right]+\frac{\mu\left[i\right]}{\varepsilon+||\boldsymbol{\varphi}\left[i\right]||^{2}}e\left[i\right]\boldsymbol{\varphi}\left[i\right],\label{eq:NKLMS}
\end{equation}
where $\mu\left[i\right]$ is the step size that should be chosen
to satisfy the constraints of the algorithm and $\varepsilon$ is
a small constant used to avoid numerical problems. Substituting \eqref{eq:NKLMS}
in \eqref{eq:posteriori error} and using the kernel trick to replace
dot products by kernel evaluations we arrive at:
\begin{equation}
\xi_{ap}\left[i\right]=e\left[i\right]-\frac{\mu\left[i\right]}{\varepsilon+\kappa\left(\boldsymbol{x}\left[i\right],\boldsymbol{x}\left[i\right]\right)}e\left[i\right]\kappa\left(\boldsymbol{x}\left[i\right],\boldsymbol{x}\left[i\right]\right)
\end{equation}

Assuming that the constant $\varepsilon$ is sufficiently small to
guarantee that $\nicefrac{\kappa\left(\boldsymbol{x}\left[i\right],\boldsymbol{x}\left[i\right]\right)}{\varepsilon+\kappa\left(\boldsymbol{x}\left[i\right],\boldsymbol{x}\left[i\right]\right)}\approx1$
and following the procedure stated in \cite{Diniz2008}we get:

\begin{equation}
\mu\left[i\right]=\begin{cases}
\begin{array}{c}
1-\frac{\gamma}{|e\left[i\right]|}\\
0
\end{array} & \begin{array}{c}
|e\left[i\right]|>\gamma\\
\mbox{otherwise}
\end{array}\end{cases}\label{eq:step size}
\end{equation}

We can compute $\boldsymbol{\omega}$ recursively as follows:

\begin{eqnarray}
\boldsymbol{\omega}\left[i+1\right] & = & \boldsymbol{\omega}\left[i-1\right]+\frac{\mu\left[i-1\right]e\left[i-1\right]}{\varepsilon+||\boldsymbol{\varphi}\left[i-1\right]||^{2}}\boldsymbol{\varphi}\left[i-1\right]\nonumber \\
 &  & +\frac{\mu\left[i\right]}{\varepsilon+||\boldsymbol{\varphi}\left[i\right]||^{2}}e\left[i\right]\boldsymbol{\varphi}\left[i\right]
\end{eqnarray}

\[
\vdots
\]
\begin{equation}
\boldsymbol{\omega}\left[i+1\right]=\boldsymbol{\omega}\left[0\right]+\sum_{k=1}^{i}\frac{\mu\left[k\right]}{\varepsilon+||\boldsymbol{\varphi}\left[k\right]||^{2}}e\left[k\right]\boldsymbol{\varphi}\left[k\right]
\end{equation}

Setting $\boldsymbol{\omega}\left[0\right]$ to zero leads to:

\begin{equation}
\boldsymbol{\omega}\left[i+1\right]=\sum_{k=1}^{i}\frac{\mu\left[k\right]}{\varepsilon+||\boldsymbol{\varphi}\left[k\right]||^{2}}e\left[k\right]\boldsymbol{\varphi}\left[k\right]
\end{equation}

The output $f(\boldsymbol{\varphi}\left[i+1\right])=\boldsymbol{\omega^{T}}\left[i+1\right]\boldsymbol{\varphi}\left[i+1\right]$of
the filter to a new input $\boldsymbol{\varphi}\left[i+1\right]$
can be computed as the following inner product
\begin{equation}
f(\boldsymbol{\varphi}\left[i+1\right])=\left[\sum_{k=1}^{i}\frac{\mu\left[k\right]e\left[k\right]}{\varepsilon+||\boldsymbol{\varphi}\left[k\right]||^{2}}\boldsymbol{\varphi}^{T}\left[k\right]\right]\boldsymbol{\varphi}\left[i+1\right]
\end{equation}

\begin{equation}
=\sum_{k=1}^{i}\frac{\mu\left[k\right]e\left[k\right]}{\varepsilon+||\boldsymbol{\varphi}\left[k\right]||^{2}}\boldsymbol{\varphi}^{T}\left[k\right]\boldsymbol{\varphi}\left[i+1\right].
\end{equation}

Using the kernel trick we obtain that the output is equal to:
\begin{equation}
\sum_{k=1}^{i}\frac{\mu\left[k\right]e\left[k\right]}{\varepsilon+\kappa\left(\boldsymbol{x}\left[k\right],\boldsymbol{x}\left[k\right]\right)}\kappa\left(\boldsymbol{x}\left[k\right]\boldsymbol{,}\boldsymbol{x}\left[i+1\right]\right),\label{eq:update equation}
\end{equation}
where $\mu\left[k\right]$ is given by \eqref{eq:step size} . Let
us define a coefficient vector $\boldsymbol{a}=\mu\left[i\right]e\left[i\right]$,
so that equation \eqref{eq:update equation} becomes:

\begin{equation}
\sum\frac{a_{i}}{\varepsilon+\kappa\left(\boldsymbol{x}\left[k\right],\boldsymbol{x}\left[k\right]\right)}\kappa\left(\boldsymbol{x}\left[k\right]\boldsymbol{,}\boldsymbol{x}\left[i+1\right]\right)\label{eq:SM-NKLMS}
\end{equation}

Equations \eqref{eq:step size} -\eqref{eq:SM-NKLMS} summarize the
algorithm proposed. We set the initial value of $\boldsymbol{\omega}$
to zero as well as the first coefficient. As new inputs arrive we
can calculate the output of the system with \eqref{eq:SM-NKLMS}.
Then the error may be computed and if it exceeds the bound established
we calculate the step size with \eqref{eq:step size}. Finally we
update the coefficients $\boldsymbol{a}_{i}$ . Note that some coefficients
may be zero due the data selective characteristic of the algorithm.
We do not need to store the zero coefficients as they do not contribute
to the output computations, resulting in a saving of resources. This
is an important result because it controls in a natural way the growing
network created by the algorithm. In stationary environments the algorithm
will limit the growing structure.

Consider now the KAP algorithm, which uses the last $K$ inputs to
update the coefficients. Based on this fact, let us redefine our problem
and use the past $K$ constraint sets to perform the update. Under
this scope it is also convenient to express the exact membership as
follows:

\begin{equation}
\psi\left[i\right]=\left(\bigcap_{j=0}^{i-K}\mathcal{H}\left[j\right]\right)\left(\bigcap_{l=i-K+1}^{i}\mathcal{H}\left[l\right]\right)=\psi^{i-K}\left[i\right]\bigcap\psi^{K}\left[i\right],
\end{equation}
where $\psi^{K}\left[i\right]$ designates the use of $K$ constraint
sets for updating. This means that the vector $\boldsymbol{\omega}\left[i\right]$
should belong to $\psi^{K}\left[i\right]$ . In order to develop the
SM-KAP algorithm we need to set several bounds $\bar{\gamma}_{k}\left[i\right]$,
for $k=1,\ldots,K$, so that the error magnitudes should satisfy this
constraints after updating. It follows that there exists a space $S\left(i-k+1\right)$
containing all vectors $\boldsymbol{\omega}$ satisfying $d\left(i-k+1\right)-\boldsymbol{\omega^{T}}\boldsymbol{\varphi}\left(i-k+1\right)=\bar{\gamma}_{k}\left[i\right]$
for $k=1,\ldots,K$. The SM-KAPA should perform an update whenever
$\boldsymbol{\omega}\left[i\right]\notin\psi^{K}\left[i\right]$,
so that the equation $\parallel\boldsymbol{\omega}\left[i\right]-\boldsymbol{\omega}\left[i-1\right]\parallel^{2}$
subject to $\boldsymbol{d}\left[i\right]-\boldsymbol{\Phi}^{T}\left[i\right]\boldsymbol{\omega}\left[i\right]=\bar{\boldsymbol{\gamma}}\left[i\right]$
should be minimized, where $\bar{\boldsymbol{\gamma}}\left[i\right]$
is a vector containing all the $K$ bounds. This constraint can also
be expressed as $\boldsymbol{d}\left[i\right]-\bar{\boldsymbol{\gamma}}\left[i\right]=\boldsymbol{\Phi}^{T}\left[i\right]\boldsymbol{\omega}\left[i\right]$.
Solving the problem with the method of the Lagrange multipliers we
get:

\begin{eqnarray}
\mathcal{L}\left(\boldsymbol{\omega}\left[i\right]\right) & = & \parallel\boldsymbol{\omega}\left[i\right]-\boldsymbol{\omega}\left[i-1\right]\parallel^{2}\nonumber \\
 &  & +\boldsymbol{\lambda}^{T}\left[i\right]\left(\boldsymbol{d}\left[i\right]-\boldsymbol{\Phi}^{T}\left[i\right]\boldsymbol{\omega}\left[i\right]-\bar{\boldsymbol{\gamma}}\left[i\right]\right),
\end{eqnarray}
where $\boldsymbol{\lambda}^{T}\left[i\right]$ is the vector of Lagrange
multipliers. Now we can compute the gradient of $\mathcal{L}\left(\boldsymbol{\omega}\left[i\right]\right)$
and equate it to zero.
\begin{equation}
\frac{\partial\mathcal{L}\left(\boldsymbol{\omega}\left[i\right]\right)=}{\partial\boldsymbol{\omega}\left[i\right]}2\boldsymbol{\omega}\left[i\right]-2\boldsymbol{\omega}\left[i-1\right]-\boldsymbol{\lambda}^{T}\left[i\right]\boldsymbol{\Phi}^{T}\left[i\right]=\mathbf{0}
\end{equation}
\begin{equation}
\boldsymbol{\omega}\left[i\right]=\boldsymbol{\omega}\left[i-1\right]+\frac{1}{2}\boldsymbol{\Phi}\left[i\right]\boldsymbol{\lambda}\left[i\right]
\end{equation}
\begin{equation}
\boldsymbol{d}\left[i\right]-\bar{\boldsymbol{\gamma}}\left[i\right]=\boldsymbol{\Phi}^{T}\left[i\right]\left(\boldsymbol{\omega}\left[i-1\right]+\frac{1}{2}\boldsymbol{\Phi}\left[i\right]\boldsymbol{\lambda}\left[i\right]\right)
\end{equation}
\begin{equation}
\boldsymbol{d}\left[i\right]-\bar{\boldsymbol{\gamma}}\left[i\right]=\boldsymbol{\Phi}^{T}\left[i\right]\boldsymbol{\omega}\left[i-1\right]+\boldsymbol{\Phi}^{T}\left[i\right]\boldsymbol{\Phi}\left[i\right]\frac{\boldsymbol{\lambda}\left[i\right]}{2}
\end{equation}
\begin{equation}
\frac{\boldsymbol{\lambda}\left[i\right]}{2}=\left(\boldsymbol{\Phi}^{T}\left[i\right]\boldsymbol{\Phi}\left[i\right]\right)^{-1}\left(\boldsymbol{e}\left[i\right]-\bar{\boldsymbol{\gamma}}\left[i\right]\right),
\end{equation}

We can now formulate the updating equation, which is used as long
as the error is greater than the established bound, i.e., $|e\left[i\right]|>\bar{\gamma}$
\begin{equation}
\boldsymbol{\omega}\left[i\right]=\boldsymbol{\omega}\left[i-1\right]+\boldsymbol{\Phi}\left[i\right]\left(\boldsymbol{\Phi}^{T}\left[i\right]\boldsymbol{\Phi}\left[i\right]\right)^{-1}\left(\boldsymbol{e}\left[i\right]-\bar{\boldsymbol{\gamma}}\left[i\right]\right),\label{eq:update SM-KAPA}
\end{equation}
where we have to consider that the vector $\boldsymbol{e}\left[i\right]$
is composed by the actual error and all $K-1$ a posteriori errors,
corresponding to the $K-1$ last inputs used. This means that vector
$\boldsymbol{e}\left[i\right]$ is expressed by $\left[\begin{array}{cccc}
e\left[i\right] & e_{ap}\left[i-1\right] & \cdots & e_{ap}\left[i-K+1\right]\end{array}\right]$ ,where $e_{ap}\left[i-k\right]$ denotes the a posteriori error computed
using the coefficients at iteration $i$. In other words, $e_{ap}\left[i-k\right]=d\left[i-k\right]-\boldsymbol{\varphi}^{T}\left[i-k\right]\boldsymbol{\omega}\left[k\right]$.

Let us now consider a simple choice for vector $\bar{\boldsymbol{\gamma}}\left[i\right]$.
We can exploit the fact that the a posteriori error was updated to
satisfy the constraint $\boldsymbol{d}\left[i\right]-\boldsymbol{\Phi}^{T}\left[i\right]\boldsymbol{\omega}\left[i\right]=\bar{\boldsymbol{\gamma}}\left[i\right]$.That
means that we can set the values of $\bar{\gamma}_{k}\left[i\right]$
equal to $e_{ap}\left[i-k+1\right]$for $i\neq1$. Substituting this
condition in equation \eqref{eq:update SM-KAPA} we obtain:
\begin{equation}
\boldsymbol{\omega}\left[i\right]=\boldsymbol{\omega}\left[i-1\right]+\boldsymbol{\Phi}\left[i\right]\left(\boldsymbol{\Phi}^{T}\left[i\right]\boldsymbol{\Phi}\left[i\right]\right)^{-1}\left(e\left[i\right]-\bar{\gamma_{1}}\left[i\right]\right)\boldsymbol{u},
\end{equation}
where $\boldsymbol{u}=\left[\begin{array}{cccc}
1 & 0 & \cdots & 0\end{array}\right]^{T}$. We can now select $\bar{\gamma_{1}}\left[i\right]$ as in the SM-NKLMS
so that

\begin{equation}
\bar{\gamma_{1}}\left[i\right]=\bar{\gamma}\frac{e\left[i\right]}{|e\left[i\right]|}
\end{equation}

\begin{equation}
\boldsymbol{\omega}\left[i\right]=\boldsymbol{\omega}\left[i-1\right]+\boldsymbol{\Phi}\left[i\right]\left(\boldsymbol{\Phi}^{T}\left[i\right]\boldsymbol{\Phi}\left[i\right]\right)^{-1}\left(\eta\left[i\right]e\left[i\right]\right)\boldsymbol{u}
\end{equation}

\begin{equation}
\eta\left[i\right]=\begin{cases}
1-\frac{\bar{\gamma}}{|e\left[i\right]|} & |e\left[i\right]|>\bar{\gamma}\\
0 & \mbox{Other Case}
\end{cases}
\end{equation}

\begin{equation}
\boldsymbol{\omega}\left[i\right]=\sum_{j=1}^{i-1}a_{j}\left[i-1\right]\boldsymbol{\varphi}\left[j\right]+\left(\eta\left[i\right]e\left[i\right]\right)\boldsymbol{\Phi}\left[i\right]\tilde{\mathbf{A}}\left[i\right],
\end{equation}
where the matrix $\tilde{\mathbf{A}}\left[i\right]$ was redefined
as
\begin{equation}
\tilde{\mathbf{A}}\left[i\right]=\left(\boldsymbol{\Phi}^{T}\left[i\right]\boldsymbol{\Phi}\left[i\right]+\epsilon\mathbf{I}\right)^{-1}\boldsymbol{u}
\end{equation}

\begin{equation}
a_{k}\left[i\right]=\begin{cases}
\eta\left[i\right]e\left[i\right]\tilde{a}_{k}\left[i\right], & k=i\\
a_{k}\left[i-1\right]+\eta\left[i\right]e\left[i\right]\tilde{a}_{K+k-i}\left[i\right], & i-K+1\leq k\\
a_{k}\left[i-1\right] & 1\leq k<i-K+1
\end{cases}
\end{equation}
 \vspace{-0.35em}

\section{Simulations}

\vspace{-0.35em}

In this section we analyze the performance of the algorithms proposed
for a time series prediction task. We used two different time series
to perform the tests, the Mackey Glass time series and a laser generated
time series. First we separate the data into two sets, one for training
and the other for testing as suggested in \cite{LiuPrincipeHaykin2010}.The
time-window was set to seven and the prediction horizon to one, so
that the last seven inputs of the time series were used to predict
the value one step ahead. Additionally, both time series were corrupted
by additive Gaussian noise with zero mean and standard deviation equal
to 0.04. The Gaussian kernel was used in all the algorithms to perform
all the experiments. Using the silver rule and after several tests,
the bandwith of the kernel was set to one.

\begin{figure}[H]
\centering{}\includegraphics[scale=0.35]{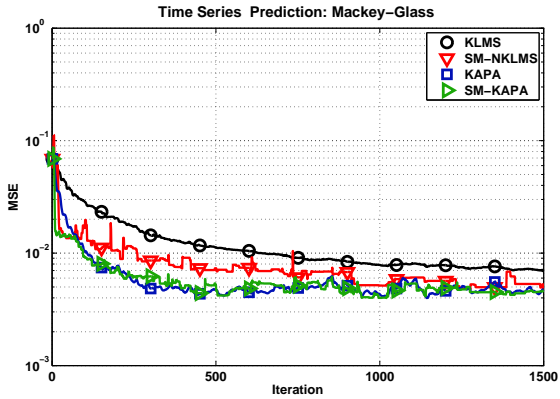}\caption{Learning Curve
of the Kernel Adaptive Algorithms for the Mackey-Glass Time Series
prediction\label{fig:Learning-Curve MG}}
\end{figure}

For the first experiment we analyze the performance of the adaptive
algorithms over the Mackey-Glass time series. A total of 1500 sample
inputs were used to generate the learning curve and the prediction
was performed over 100 test samples. For the KAPA and the SM-KAPA
algorithms, $K$ was set to $7$ so that the algorithms used the last
seven input samples as a single input. For the KLMS algorithm the
step size was set to $0.05$.The error bound for the SM-NKLMS and
the SM-KAPA algorithm was set to $\sqrt{5}\sigma$. The final results
of the algorithms tested are shown in table \ref{tab:Performance MG time series}
where the last 100 data points of each learning curve were averaged
to obtain the MSE. The learning curves of the algorithms based on
kernels is presented in Figure \ref{fig:Learning-Curve MG}. From
the curves, we see that the algorithms proposed outperform conventional
algorithms in convergence speed.

\begin{table}[H]
\begin{centering}
\caption{Performance comparision on Mackey-Glass time series prediction\label{tab:Performance MG time series}}

\par\end{centering}

\centering\resizebox{7cm}{1.7cm}{

\begin{centering}
\begin{tabular}{|c|c|c|}
\hline
\textbf{Algorithm} & \textbf{Test MSE} & \textbf{Standard} \textbf{Deviation}\tabularnewline
\hline
\hline
LMS & 0.0230680 & +/-0.00020388\tabularnewline
\hline
NLMS & 0.0213180 & +/-0.00017318\tabularnewline
\hline
SM-NLMS & 0.0202340 & +/-0.00084243\tabularnewline
\hline
APA & 0.0208600 & +/-0.00231500\tabularnewline
\hline
SM-APA & 0.0204340 & +/-0.00228940\tabularnewline
\hline
KLMS & 0.0075596 & +/-0.00030344\tabularnewline
\hline
SM-NKLMS & 0.0054699 & +/-0.00046209\tabularnewline
\hline
KAPA2 & 0.0047812 & +/-0.00041816\tabularnewline
\hline
\multirow{1}{*}{SM-KAPA} & 0.0046603 & +/-0.00032855\tabularnewline
\hline
\end{tabular}
\par\end{centering}

}
\end{table}

\begin{figure}[H]
\begin{centering}
\includegraphics[scale=0.35]{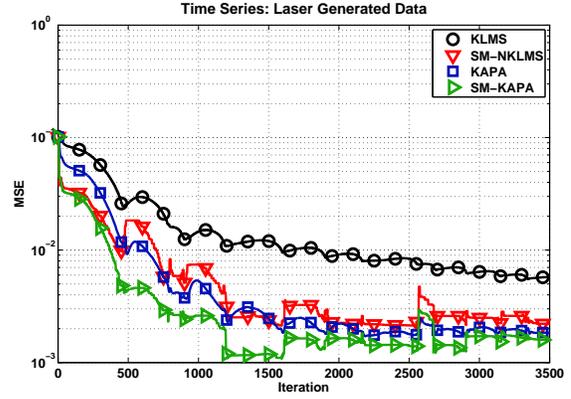}
\par\end{centering}

\caption{Learning Curve of the SM-KAPA for the L\label{fig:Learning-Curve Laser}aser
Time Series prediction}
\end{figure}

In the second experiment we consider the performance of the proposed
algorithms over a laser generated time series. In this case , 3500
sample inputs were used to generate the learning curve and the prediction
was performed over 100 test samples. The setup used in the previous
experiment was considered. Table \ref{tab:Performance Laser time series}
summarizes the MSE obtained for every algorithm tested. The final
learning curves are showed in Figure \ref{fig:Learning-Curve Laser}.

\begin{table}[H]
\begin{centering}
\caption{Performance\label{tab:Performance Laser time series} comparision
on laser generated time series prediction}

\par\end{centering}

\centering\resizebox{7cm}{1.8cm}{

\begin{centering}
\begin{tabular}{|c|c|c|}
\hline
\textbf{Algorithm} & \textbf{Test MSE} & \textbf{Standard} \textbf{Deviation}\tabularnewline
\hline
\hline
LMS & 0.0214290 & +/-0.00035874\tabularnewline
\hline
NLMS & 0.0197260 & +/-0.00101250\tabularnewline
\hline
SM-NLMS & 0.0246950 & +/-0.00647190\tabularnewline
\hline
APA & 0.0255460 & +/-0.00465890\tabularnewline
\hline
SM-APA & 0.0200020 & +/-0.00154490\tabularnewline
\hline
KLMS & 0.0090129 & +/-0.00067428\tabularnewline
\hline
SM-NKLMS & 0.0038472 & +/-0.00054237\tabularnewline
\hline
KAPA2 & 0.0028253 & +/-0.00030613\tabularnewline
\hline
\multirow{1}{*}{SM-KAPA} & 0.0029454 & +/-0.00019424\tabularnewline
\hline
\end{tabular}
\par\end{centering}

}
\end{table}

\begin{figure}[H]
\begin{centering}
\includegraphics[scale=0.35]{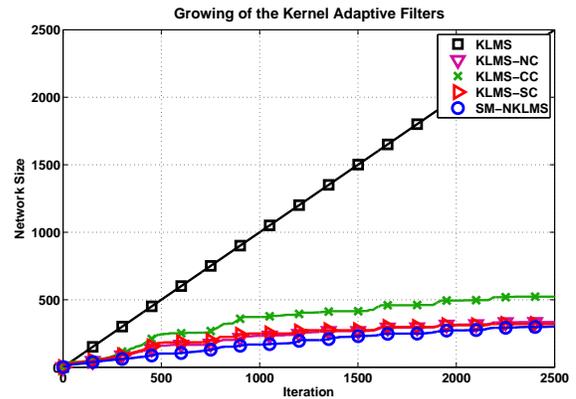}
\par\end{centering}
\caption{Dictionary \label{fig:Dictionary-Size}Size vs Iterations}
\end{figure}

On the next experiment we study the size of the dictionary generated
by the conventional KLMS algorithm and by the proposed SM-NKLMS algorithm.
The result is presented in Figure \ref{fig:Dictionary-Size}. We see
that the proposed algorithm naturaly limits the size of the dictionary.

\begin{figure}[H]
\centering{}\includegraphics[scale=0.35]{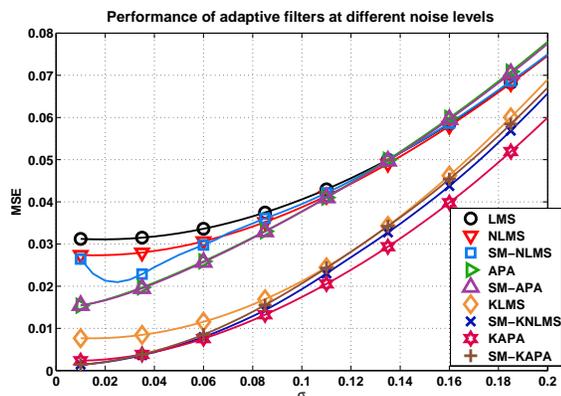}\caption{Robustness\label{fig:Robustness}}
\end{figure}

As a final experiment, we analyze and compare the robustness of the
algorithm proposed with respect to the conventional algorithms. Figure
\eqref{fig:Robustness} shows the results obtained. It is clear that
the SM-NKLMS exhibits a better perfomance than the KLMS algorithm.
In general, all kernel algorithms overperform their linear counterparts.

\vspace{-0.35em}

\section{Conclusions}

\vspace{-0.35em}

In this paper two data selective kernel-type algorithms were
presented, the SM-KNLMS and the SM-KAP algorithms. Both algorithms
have a faster convergence speed than the conventional algorithms.
They also have the advantage of naturally limiting the size of the
dictionary created by kernel-based algorithms and a good noise
robustness. In general, the proposed algorithms outperform the
existing kernel-based algorithms. \vspace{-0.35em}

\nonumber
{\footnotesize{}\bibliographystyle{plain}
\bibliography{adaptivefilters}
}{\footnotesize \par}
\end{document}